\documentclass[10pt,twocolumn,letterpaper]{article}

\usepackage[pagenumbers]{wacv} 

\usepackage{graphicx}
\usepackage{amsmath}
\usepackage{amssymb}
\usepackage{booktabs}
\usepackage{tabularx}
\usepackage{float}
\usepackage{balance}
\usepackage[accsupp]{axessibility}  

\usepackage[pagebackref,breaklinks,colorlinks]{hyperref}

\usepackage[capitalize]{cleveref}
\crefname{section}{Sec.}{Secs.}
\Crefname{section}{Section}{Sections}
\Crefname{table}{Table}{Tables}
\crefname{table}{Tab.}{Tabs.}

\def\wacvPaperID{2110} 
\def\confName{WACV}
\def\confYear{2024}

\begin{document}

\title{NOMAD: A Natural, Occluded, Multi-scale Aerial Dataset, \\for Emergency Response Scenarios}

\author{Arturo Miguel Russell Bernal, Walter Scheirer, Jane Cleland-Huang\\
Computer Science and Engineering Department, University of Notre Dame, Indiana, USA\\
{\tt\small arussel8@nd.edu, walter.scheirer@nd.edu, janehuang@nd.edu}
}
\maketitle

\begin{abstract}
\vspace{-0.2cm}
   With the increasing reliance on small Unmanned Aerial Systems (sUAS) for Emergency Response Scenarios, such as Search and Rescue, the integration of computer vision capabilities has become a key factor in mission success. Nevertheless, computer vision performance for detecting  humans severely degrades when shifting from ground to aerial views.  Several aerial datasets have been created to mitigate this problem, however, none of them has specifically addressed the issue of occlusion, a critical component in Emergency Response Scenarios.  Natural, Occluded, Multi-scale Aerial Dataset (NOMAD) presents a benchmark for human detection under occluded aerial views, with five different aerial distances and rich imagery variance.  NOMAD is composed of 100 different Actors, all performing sequences of walking, laying and hiding. It includes 42,825 frames, extracted from 5.4k resolution videos, and manually annotated with a bounding box and a  label describing 10 different visibility levels, categorized according to the percentage of the human body visible inside the bounding box. This allows computer vision models to be evaluated on their detection performance across different ranges of occlusion. NOMAD is designed to improve the effectiveness of aerial search and rescue and to enhance collaboration between sUAS and humans, by providing a new benchmark dataset for human detection under occluded aerial views.  Full dataset can be found at: \href{https://github.com/ArtRuss/NOMAD}{https://github.com/ArtRuss/NOMAD}.
\end{abstract}
\vspace{-0.5cm}

\section{Introduction}
\label{sec:intro}

Advances in technology, including improvements in edge computing and Artificial Intelligence (AI), have led to increased use of small Unmanned Aerial Systems (sUAS) across a broad range of applications \cite{UAV_marketForecast, survey_civilApplications, survey_edgeAI_UAVs}, such as emergency response \cite{survey_civilApplications, survey_edgeAI_UAVs, UAV_SARMap, bernal2023hierarchically}. sUAS are empowered to perform Computer Vision (CV) tasks, such as  aerial surveillance and autonomous person detection and tracking, where timely and efficient performance potentially can make the difference between life or death \cite{news_drowning, news_hikingAfrica, news_lostWoods}. Higher levels of sUAS autonomy, supported by CV, increase collaboration between humans and sUAS, allowing emergency responders to focus attention on mission level goals \cite{chi20-partnerships, DBLP:conf/seams/Cleland-HuangAV22} while sUAS perform lower-level person detection tasks.  

However, there are many open challenges in deploying CV on sUAS for emergency response \cite{challenges_tawfiq, challenges_murphy}. These challenges include the non-trivial, highly prevalent problem of occlusion, which occurs when targets of aerial search are  partially hidden from view. For example, a drowning victim who is partially submerged in water, people buried in debris following an earthquake, hidden by smoke in a fire, or laying behind trees and rocks in search and rescue missions. Occlusion could also be intentional when a suspect is hiding from law-enforcement, caused by pose and image perspective, or introduced in far-distance aerial views due to glare, shades, blur, and low resolution. Prior CV research on occlusion has focused on generic object detection \cite{survey_generic, survey_generic_occlusion}, as well as on pedestrian detection \cite{survey_pedestrian_occlusion}, demonstrating how occlusion drastically affects model performance \cite{survey_pedestrian_occlusion, semantic_occlusion, occlusionvshumans}. However, the occlusion problem is exacerbated even further when shifting from ground to aerial views \cite{survey_aerial}, where additional challenges surrounding the incorporation of CV capable sUAS into emergency response scenarios include biased training datasets, coupled with real-life challenges such as vibration, harsh weather and low visibility conditions, diverse scenery, and the need for generalization at different distances and resolutions.  CV systems deployed for emergency response must be able to reliably handle person detection under all of these variable conditions. 

\begin{figure*}[t]
  \centering
   \includegraphics[trim=0cm 0.5cm 0cm 0.5cm,clip,width=0.95\linewidth]{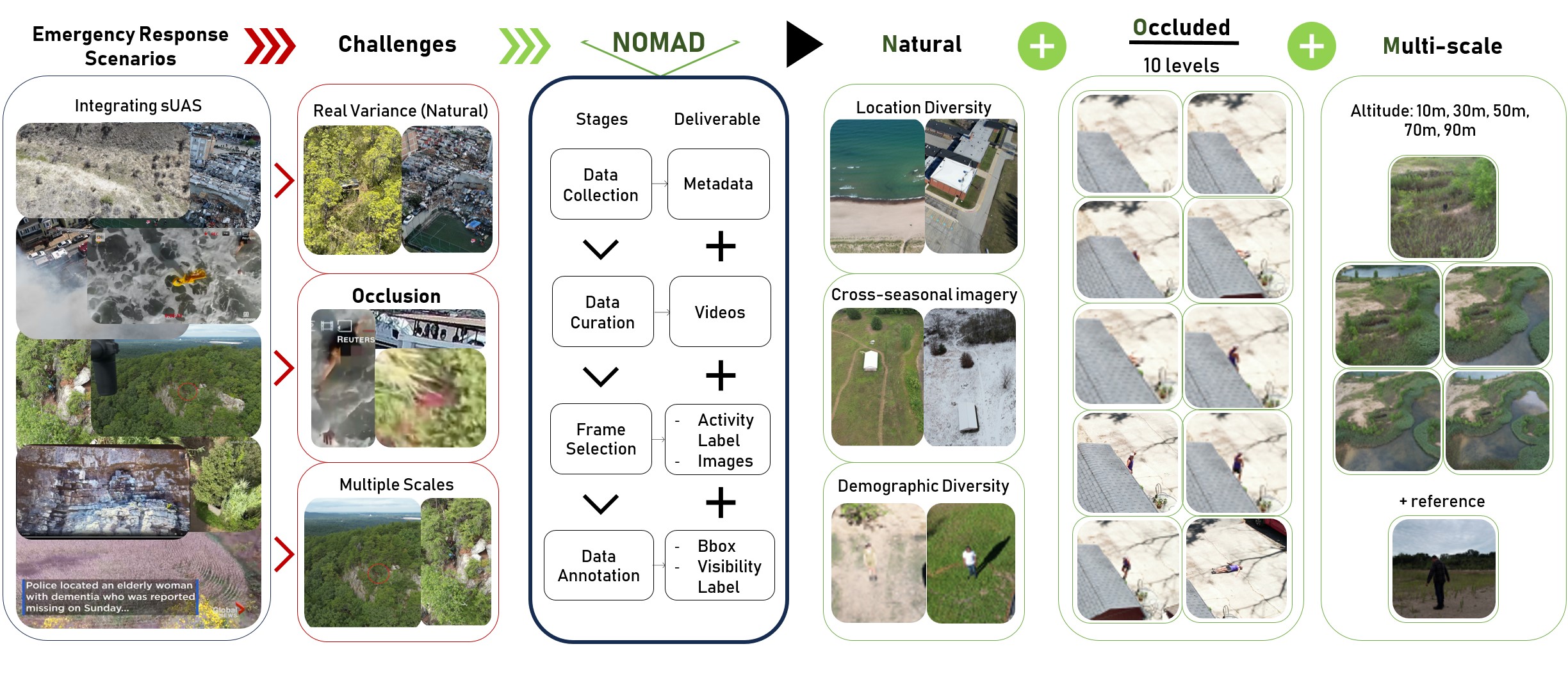}\vspace{-0.2cm}
   \caption{Development and characteristics of NOMAD.  Integration of sUAS into emergency response scenarios have aided first responders and rescued victims \cite{news_drowning, news_hikingAfrica, news_lostWoods, news_earthquake, news_girl, news_fire, news_teamwork, news_oldadult, news_firehomes} (first column). Nevertheless, multiple challenges inherent to these situations degrade CV performance and halt sUAS full integration, including the highly prevalent problem of occlusion (second column). We present NOMAD, Natural Occluded Multi-scale Aerial Dataset, providing the research community with emergency response related videos and selected frames, as well as rich metadata and annotations, including a visibility label (third column). Facing emergency response scenarios, key characteristics of our dataset are: \textit{Natural}: diversity of filming locations, cross-seasonal imagery, including winter scenarios, and a demographic diversity on gender, age and race, ranging from 18 to 78 years old, and including White Caucasians, Latinos, African descent, Asians, South Asians, Middle Eastern and Pacific Islander; \textit{Occluded}: 10 defined ranges of occlusion, with a visibility label assigned to every bounding box; \textit{Multi-scale}: five different distances, ranging from 10m to 90m altitude, and a ground reference view for every actor.}  \vspace{-0.5cm}
   \label{fig:NOMAD}
\end{figure*}

 We therefore address these challenges through presenting NOMAD (Natural Occluded Multi-scale Aerial Dataset), a benchmark dataset aimed at human detection under occluded aerial views, as summarized in \cref{fig:NOMAD}. NOMAD is composed of 100 different actors, each performing sequences of walking, laying and hiding. It includes 42,825 frames, extracted from 5.4k resolution videos. Actors are manually annotated with a bounding box and a label describing 10 different visibility levels, categorized according to the percentage of the human body visible inside the bounding box, allowing the detection performance of CV models to be evaluated across 10 different ranges of occlusion. \Cref{fig:NOMAD} summarizes the key characteristics of our dataset, including: \textit{Natural}: representing a variety of natural and man-made locations; cross-seasonal imagery, ranging from summer to winter scenarios; demographic variety on age and race, ranging from 18 to 78 years old, and including White Caucasians, Latinos, African descent, Asians, South Asians, Middle Eastern and Pacific Islander; \textit{Occluded}: with routines created to include occlusion and a visibility label assigned to every bounding box annotated; \textit{Multi-scale}:  with five different distances, ranging from 10m to 90m altitude, and a ground reference view for every actor. 

The remainder of this article is organized as follows. \Cref{sec:RelatedWork} presents related work. 
\Cref{sec:DataCollection} describes the data collection process. \Cref{sec:DataAnnotation} describes the data curation, key-frame selection and data annotation. \Cref{sec:DatasetCharacteristics} discusses NOMAD characteristics and its potential uses.  \Cref{sec:CVmodels} reports baseline results achieved using state-of-the-art CV detection models under different levels of occlusion, and \cref{sec:FutureWorkConclusions} summarizes the contributions of the work.
\vspace{-0.15cm}

\section{Related Work}
\label{sec:RelatedWork}

\subsection{Mobile Robotics for Emergency Response}
There are numerous challenges associated with integrating  mobile robotics into emergency response missions  \cite{challenges_tawfiq, challenges_murphy, survey_civilApplications, survey_edgeAI_UAVs, challenges_walter, nagatani}. 
Researchers, focusing on ground mobile robots, have explored mapping of emergency scenes \cite{ground_map_reinforcementLearning, ground_3dmap, ground_SMURF}, improved communication networks \cite{ground_network}, and specialized architectures \cite{ground_gasNose, ground_snake}. User studies have demonstrated the benefits of including aerial robots in emergency response \cite{usability}, potentially working in collaboration with ground robots \cite{aerial_ground_collaboration}, to enhance surveying and mapping capabilities \cite{aerial_helicopters}. Other studies have  explored the integration of additional sensors, such as ground penetrating radar \cite{aerial_groundpenetratingradar}, or cellphone tracking for missing person search \cite{aerial_trackingSignal}. Finally, several researchers have explored efficient collaborations between humans and sUAS at the intersection of software engineering and human computer interaction \cite{hci_murphy, chi20-partnerships, DBLP:conf/seams/Cleland-HuangAV22, droneresponse}.

\subsection{Real-World Object Detection}
There are numerous challenges related to utilizing aerial CV for real-time emergency response \cite{survey_generic}.  Real-time CV applications tend to leverage the latest versions of the YOLO family \cite{yolov7, yolov8_ultralytics}, as well as their modifications \cite{driving_yolo, edge_yolo_wacv, pedestrian_yolo, object_afc_edge_yolo}, while other methods explore attention for object detection \cite{attention} and multimodal techniques \cite{pedestrian_multimodal}. The most recent work has focused on incremental learning of unknown classes, in the modality known as Open World Object Detection \cite{OWOD_CVPR_birth, OWOD_CVPR_model, OWOD_PROB, OWOD_revisiting}, as well as its variations \cite{UCOWOD, OSODD}.  The challenges of object detection under occlusion have also been studied \cite{survey_generic_occlusion, survey_pedestrian_occlusion, occlusion_oldsurvey}. Finally, techniques incorporating human perception have been explored for object detection \cite{humanperception_objectdetection} and other machine learning tasks \cite{Walter_original, walter_brain, sam_writting, justin_tl, adam_iris, adam_new, cyborg}, demonstrating a plausible approach to handling occlusion \cite{occlusion_humanperception, psyphy, justin_guiding,occlusionvshumans}.

\subsection{Aerial Datasets}
While many datasets have been collected to aid aerial detection of humans in search and rescue (SAR), none of them have addressed the critical issue of occlusion. HERIDAL \cite{heridal} comprises of approximately 500 labelled 4,000 by 3,000 pixel images suitable for object detection tasks. SARD \cite{sard} comprises 1,981 manually labeled images extracted from video frames of persons simulating search and rescue situations in roads, quarries, grass land, and forested areas, under diverse weather conditions.  However,  both datasets lack rich generalization characteristics and environmental diversity. The recently published  WiSARD \cite{WiSARD} dataset, comprises the richest set of images associated with wilderness SAR scenarios, with 33,786 labeled RGB images, 22,156 labeled thermal images, and a subset consisting of 15,453 temporally synchronized visual-thermal image pairs. In addition to the useful multimodal imagery, the dataset includes environmental diversity across seasons and times of the day and night. WiSARD represents the richest dataset for \textit{blind search} in wilderness scenarios, that is, search for any person on an area rather than the search for an specific person; NOMAD provides richer demographic diversity, includes man-made scenarios, provides rich metadata of actors, controlled multi-scales, and provides a new benchmark for occlusion. It is the only dataset, to our knowledge, to systematically address the issue of occlusion. 

The BIRDSAI, VisDrone and UAVDT datasets \cite{visdrone, birdsai, uavdt} incorporate occlusion labels into their annotations; however, they lack rich human metadata. BIRDSAI is a long-wave thermal infrared dataset containing nighttime images of animals and humans in Southern Africa. While suitable for improving \textit{blind search} of persons in emergency scenarios, it only provides two levels of occlusion and lacks person metadata.  VisDrone consists of 288 video clips formed by 261,908 frames and 10,209 static images, and is captured by various drone-mounted cameras, covering diverse locations, environments, objects (pedestrian, vehicles, bicycles, etc.), and density. However, it provides only three levels of occlusion and also lacks person metadata.  Finally, while UAVDT provides four levels of occlusion, it focuses purely on vehicles and not people.

BRIAR, MEVID, UAV-Human, P-DESTRE and PRAI-1581 \cite{briar, mevid, uavhuman, pdestre, prai1581} provide rich metadata and are well suited for person re-identification.  BRIAR and MEVID datasets offer great diversity of camera views, with BRIAR providing long range imagery of up to 1000m.  BRIAR, so far, includes more than 350,000 still images and over 1,300 hours of video footage of approximately 1,000 subjects; MEVID, is part of the very-large-scale MEVA person activities dataset \cite{meva} and comprises 158 unique people wearing 598 outfits collected from 33 camera views. UAV-HUMAN includes 67,428 annotated video sequences of 119 subjects for action recognition, 22,476 annotated frames for pose estimation, 41,290 annotated frames of 1,144 identities for person re-identification, and 22,263 annotated frames for attribute recognition. While these three datasets represent the most complete datasets for their given purposes,  none of them reference occlusion and all lack representation of emergency response scenarios. Finally, PRAI-1581 provides 1,581 identities and P-DESTRE provides rich metadata for 269 different identities, however, filming distances are only up to 60m and 6.7m, respectively.  Additional categorized aerial datasets can be found in \cite{survey_aerial}.

Overall, NOMAD provides the demographic and environmental diversity needed to tackle the person detection task of emergency response scenarios from aerial views, while being the first dataset to include an occlusion metric for person detection, and to provide detailed metadata and controlled multi-scale, making it suitable for many other CV tasks as described in \cref{sec:DatasetCharacteristics}.

\section{Data Collection Process}
\label{sec:DataCollection}
Our data collection process followed our IRB approved protocol 21-11-6913.  In a preliminary pilot study, our data collection procedure included strict instructions regarding the percentage of the body that the actor should expose to the sUAS' camera at each step. However,  we observed that these instructions were difficult to follow causing disconnected movements, and so we replaced the instructions with simpler ones that led to more natural behavior.

\subsection{Recruitment}
As per our IRB protocol, all participants were at least 18 years old. Further, as the recruitment process evolved, participants from already well represented demographic groups were excluded in order to achieve a balanced gender distribution, a variety of age ranges, and a rich race distribution.

\subsection{Location Selection}
Approval for use of premises was attained from owners and responsible agencies for all locations filmed in the dataset. The 12 locations included: 3 different Schools, 2 paintball courts, 1 forest park, 1 golf course, 1 lake shore, 1 quarry, 2 farms, and 1 AMA flying field. This resulted in diversity of locations, including both natural and man-made influenced, and provided a variety of different types of obstacles for occlusion purposes. 

\subsection{Filming Sessions}
All filming sessions followed IRB protocol guidelines with participants being informed of the purpose of their performance, the activities to be completed, and consent forms being signed. All flights were conducted by a certified FAA Part 107 remote pilot, and all FAA protocols  were followed, with air space reserved through LAANC systems such as AirMap and DroneUp. Although efforts were made to isolate the selected locations during the filming sessions, unexpected persons appeared during a few of the filming sessions.  In most cases we stalled the filming until the person exited the scene; however, in a few cases, these persons agreed to appear on the dataset, signing a consent form. From here on, we call actors the participants performing the designated routine, while non-actors are other participants who agreed to appear in the dataset but were otherwise not engaged in the study.

Once the study introduction was completed, each actor  was assigned a unique \textit{obstacle} at the filming location, and then given instructions for performing the standard routine with respect to their obstacle as follows: \vspace{-4pt}
\begin{itemize}
    \itemsep=-.3em
    \item Starting Frame: With few exceptions, the first frame represented a view of the actor completely visible.
    \item Hiding: All actors were instructed to hide behind their obstacle two times, with small variations in their hiding trajectory.  This step allowed us to obtain varying degrees of occluded aerial views.  
    \item Laying: To provide a variety of poses, actors were asked to lay down when completely visible and when partially occluded by their obstacle.
    \item Walking: Finally, actors performed a small walking trajectory at the end of their routine.
    \item General instructions:  actors were informed of the dataset's focus on emergency response scenarios, and were therefore asked to position themselves as if they were hiding, trying to be rescued, or in need of help.
\end{itemize}

\begin{figure}[t]
  \centering
   \includegraphics[width=0.7\linewidth]{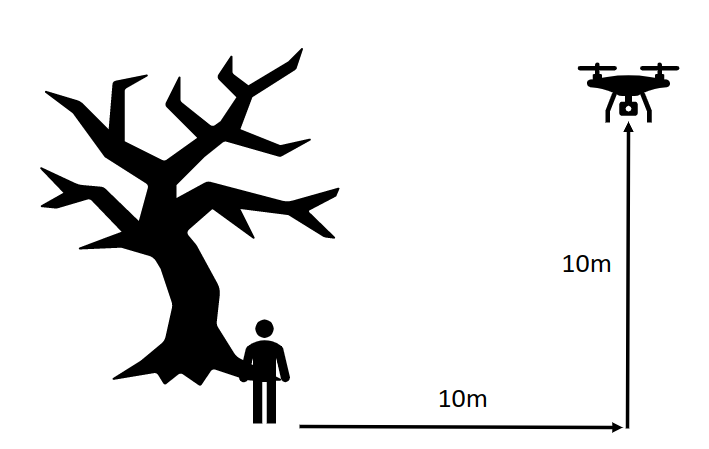}
   \caption{Filming process. Sample positioning of the sUAS at 10m horizontally and vertically from the actor's starting location.}
   \label{fig:DronePos}
\end{figure}

For the water routines, where the primary occlusion source was the water itself, small but important variations in instructions were given to simulate  various drowning scenarios. All actors were asked to repeat their routine five times, with the sUAS set at five progressively distant locations set to 10m, 30m, 50m, 70m, and 90m, with the distance measured, through the sUAS' feedback from the First Person View (FPV) screen, horizontally and vertically from the expected starting point of the actor. \Cref{fig:DronePos} illustrates the sUAS position at a distance of 10m.

Additionally, a reference view of the actor was filmed, with the sUAS positioned a few meters in front of the actor, while the actor performed 360\textdegree  rotations. The first rotation involved arms hanging down and the second with arms extended up, providing multiple views of the actor at ground level.  Finally, true negatives were also filmed by asking the actor to locate himself/herself outside of the camera view; please note that in a few cases true negatives may still contain consented non-actor participants. At the conclusion of the session, actors were given a 20\$USD prepaid card. 

\vspace{-0.5cm}
\section{Data Annotation Process}
\label{sec:DataAnnotation}

\subsection{Data curation}
Although efforts were made to avoid filming non-participants, during the revision of the films, unexpected persons were observed on a couple of videos.  For videos where the non-participant was only visible at the beginning, at the end, or at non-keyframes, trimming the video was a direct solution, representing no impact to the quality of the data. Nevertheless, situations where found where trimming the portion of the video where the non-participant appeared on screen would represent a loss of information of the actor's performance; these situations were solved by blacking out the non-participant area of the frames. 

\subsection{Metadata}
Metadata provided in this dataset can be divided into Environmental and Demographic categories, including outfit descriptions that could aid in re-identification tasks. Full list of metadata can be found in Tab. S1 in the Supplemental Material. Insight into selecting metadata factors was obtained through a previous series of semi-structured interviews with emergency responders, under IRB protocol 19-04-5269, to determine search terms used for describing missing persons. Clothing descriptions may include up to five words for salient figures. Hair length uses the same metric for males and females, with \textit{bald} meaning absence of noticeable hair, \textit{short} meaning ear-length, \textit{medium} ranging from ear- to shoulder-length, and \textit{long} meaning longer than shoulder-length.  The Location descriptor School (Nature) aggregates filming sessions where the researcher should expect nature domination despite it being filmed at a school premises. The reported weather information was obtained from the nearest weather station to the filming location. Finally, the Exposure Value (EV) was separated from the Video descriptor; while the Video descriptor is a constant for all actors, the EV parameter was found to be different than 0 on a couple of films, indicating a change in the illumination, which is a relevant parameter for computer vision tasks \cite{wu2023face_light}. Table S2 from the Supplemental Material displays the predefined lists of available colors for describing clothing and hair color. Clothing colors were selected to include the most common colors across the hue range on the HSV color space; colors for the Hair descriptor were selected based on emergency responders' classifiers.

\subsection{Keyframe selection}
Manually labeling every frame from the films would have been infeasible, therefore we selected 85 keyframes at each of the five distances for every actor. This resulted in 425 keyframes per actor, with the exception of Actor001, for whom 750 keyframes were selected. Keyframe selection was performed in accordance to the following guidelines: (1) the 85 selected frames tracked the actor across their entire routine, (2) each starting and ending frame of a trajectory was selected as a keyframe, (3) each change in direction of the actor's trajectory generated a new keyframe, (4) the set of keyframes included different poses (e.g., standing, sitting, laying down), and finally, (5) all keyframes had at least a part of the actor visible. 

\begin{figure}[t]
   \centering
   \begin{subfigure}{0.5\linewidth}
       \centering
       \includegraphics[width=\linewidth]{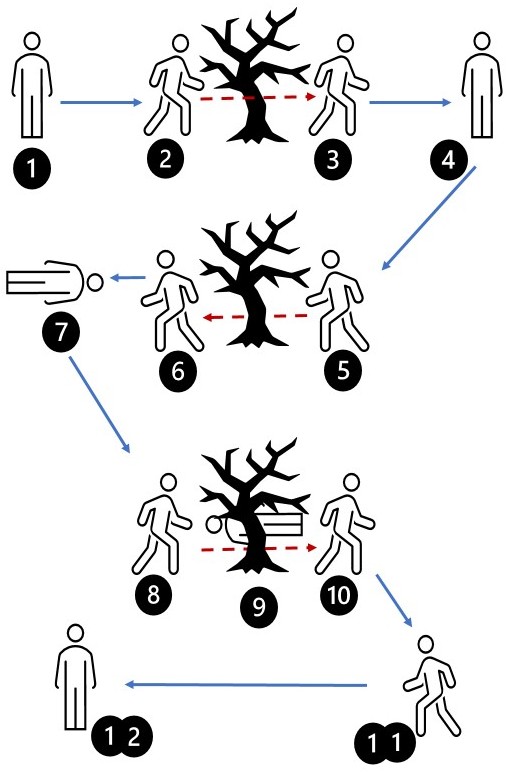}
       \caption{}
       \label{fig:keyframes}
   \end{subfigure}
   \begin{subfigure}{0.45\linewidth}
       \centering 
       \small
       \begin{tabular}{ll}
         \toprule
         Keyframes & Activity \\
         \midrule
         1 - 2 & Walking\\
         2 - 3 & Hiding\\
         3 - 5 & Walking\\
         5 - 6 & Hiding\\
         6 - 7 & Walking\\
         7 & Laying\\
         7 - 8 & Walking\\
         8 - 10 & Hiding (L)\\
         10 - 12 & Walking\\
        \bottomrule
      \end{tabular}\vspace{+0.9cm}
      \caption{}
      \label{tab:activities}
   \end{subfigure}\vspace{-0.2cm}
\caption{Keyframe selection process. (a) Sample routine with 12 keyframes selected.  Sampling for occluded views is indicated by red-dashed arrows. 
         (b) Sample Activity labels for the keyframes illustrated. Hiding (L) represents Hiding (Laying).}
\label{keyframesandactivities}
\end{figure}

Finally, when the actor interacts with an obstacle, a custom sampling is performed to obtain views with different levels of occlusion as the actor moves behind and away from an obstacle. \Cref{fig:keyframes} illustrates a sample routine with 12 keypoints selected following these guidelines. The red-dashed arrows indicate where sampling for occluded views would be performed. Activity labels were added to each keyframe as: Walking, Laying, Hiding, Hiding (Laying), Swimming, Drowning.  \Cref{tab:activities} shows sample labels for the 12 keypoints illustrated on \cref{fig:keyframes}.

\subsection{Annotations}
All 42,825 selected frames were sent to Labelbox \cite{labelbox}, a labelling company who employed expert annotators to add bounding boxes and visibility labels to all images.  

\subsubsection{Occlusion Label}
\begin{figure}[t]
  \centering
   \includegraphics[trim=0cm 1.75cm 0cm 1.75cm, clip, width=0.35\linewidth]{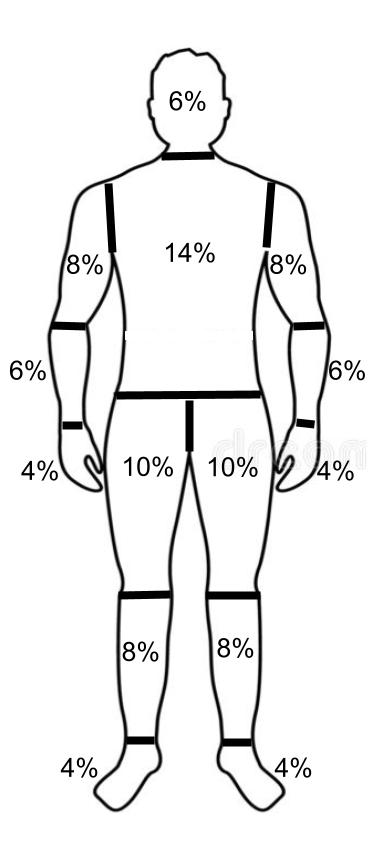}
   \caption{Visibility label calculation. Percentages assigned to each body part of a person.}\vspace{-0.4cm}
   \label{fig:BodyPercent}
\end{figure}

\Cref{fig:BodyPercent} displays how percentages were assigned to each body part of a person.  The following is the procedure used to calculate the visibility label, that is, the amount of an actor that was visible at a particular instant: (1) Given an image, identify the body parts of a person that are visible. (2) Review the percentages of the identified body parts based on \cref{fig:BodyPercent}. (3) If less than half of a body part is visible, assign half of the percentage indicated in \cref{fig:BodyPercent}. (4) If more than half of the body part is visible, assign the full percentage as indicated in \cref{fig:BodyPercent}. (5) Add up the percentages obtained from each body part. (6) Assign the sum to one of the ten ranges of visibility, with upper bounds of 10 to 100. For example, selecting 10 means that the sum obtained from the percentages is greater than zero but less than or equal to 10, while selecting 20 means that the sum is greater than 10 and less than or equal to 20, and so on.

Please note that shadows were not considered to be part of the human, and under normal circumstances, the actor’s own clothing are not treated as a source of occlusion.  Also note that although we are reporting a visibility metric, this is just the inverse of the occluded amount of the actor's body.

\section{Dataset Characteristics}
\label{sec:DatasetCharacteristics}
NOMAD provides 500 routine videos of 100 different actors, with each actor performing at five different distances, set as 10m, 30m, 50m, 70m, and 90m, including 42,825 frames manually annotated with a bounding box and visibility label. Videos' duration ranges from 30s to 180s depending on the actor's pace. It also provides one reference video per actor and 500 true negative short-duration videos, with each true negative video corresponding to one routine video.  All videos are of 30fps, MP4-H265 coding, 5.4k video quality, and frames of 5472 by 3078 pixels. 

\subsection{Natural}
\begin{figure}[t]
  \centering
   \includegraphics[width=0.9\linewidth]{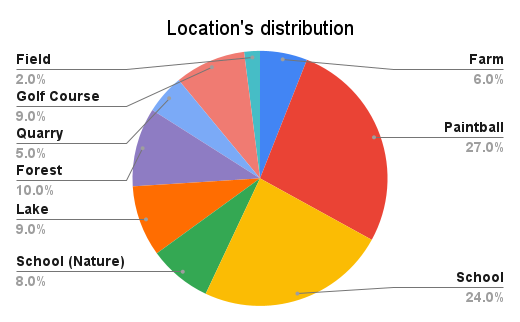}
   \caption{Distribution of the filming locations for the 100 actors.}\vspace{-0.4cm}
   \label{fig:locations}
\end{figure}

\Cref{fig:locations} shows the distribution of the 100 actors with respect to their filming locations.  The variety of locations provides coverage of natural and man-made environments, and is aimed at training CV models for effectively supporting a wide range of emergency response scenarios.

\begin{figure*}
  \centering
  \begin{subfigure}{.45\linewidth}
    \includegraphics[width=0.9\linewidth]{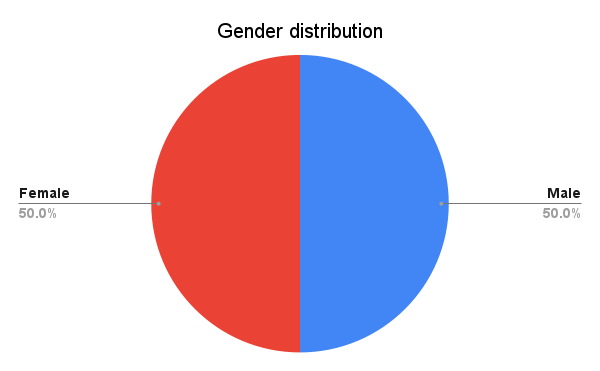}
    \caption{}
    \label{fig:dem-a}
  \end{subfigure}
  \begin{subfigure}{.45\linewidth}
    \includegraphics[width=0.9\linewidth]{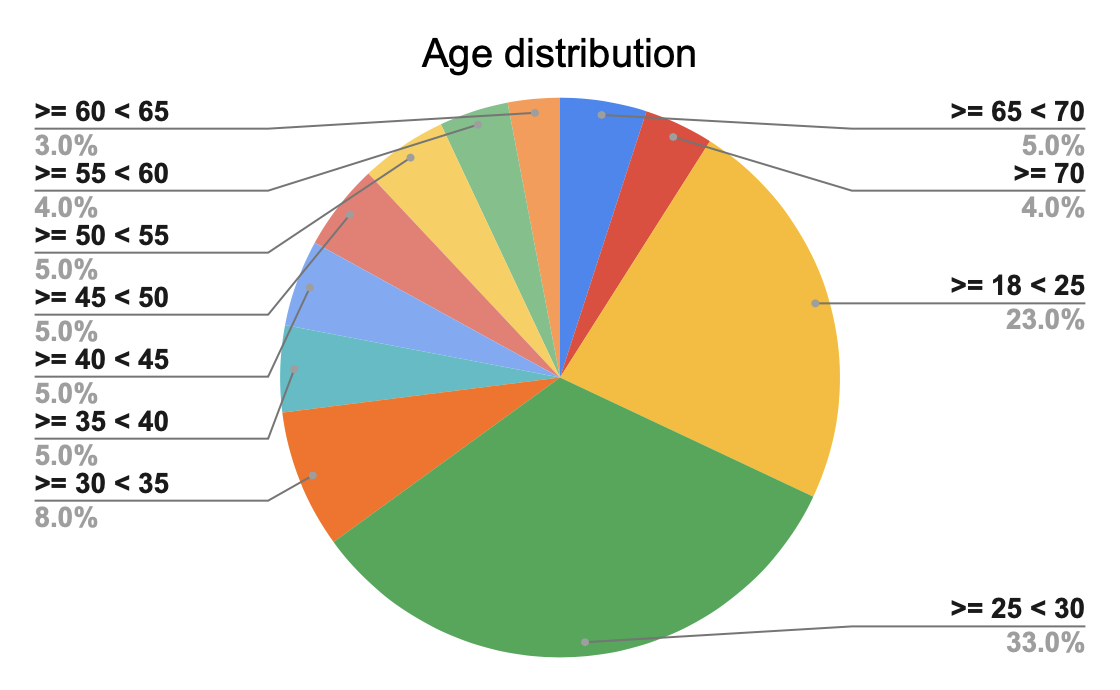}
    \caption{}
    \label{fig:dem-b}
  \end{subfigure}
  \begin{subfigure}{.45\linewidth}
    \includegraphics[width=0.9\linewidth]{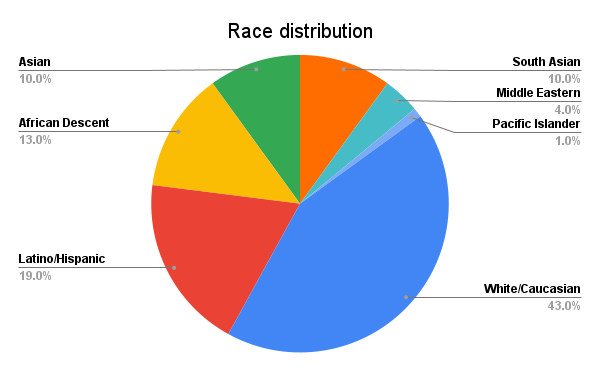}
    \caption{}
    \label{fig:dem-c}
  \end{subfigure}
  \begin{subfigure}{.45\linewidth}
    \includegraphics[width=0.9\linewidth]{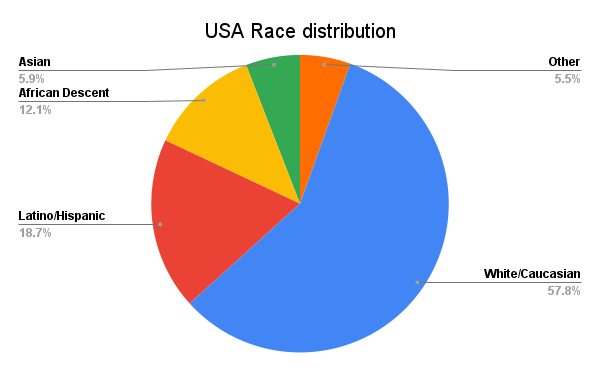}
    \caption{}
    \label{fig:dem-d}
  \end{subfigure}
  \caption{Distribution of the demographic descriptors of (a) Gender, (b) Age, and (c) Race, for our 100 actors.  Our race distribution is compared to the (d) USA race distribution, improving generalization by mitigating possible biases. }
  \label{fig:demographic}\vspace{-0.4cm}
\end{figure*}

To further increase robustness of our dataset in terms of environmental conditions, cross-seasonal imagery was collected, with temperatures ranging from 30F to 90F, wind speeds of 0MPH to 20MPH, morning, afternoon, and evening sessions, capturing hot sunny summer days, autumn colorful scenes, and winter's snowy conditions.  Finally, we made every effort to mitigate potential demographic bias to support fair and equitable emergency response.   \Cref{fig:demographic} presents the distributions of gender, age and race.  The gender distribution shows a 50/50 male/female distribution, and although age distribution shows that the majority of the population was younger than 30 years old, actors across the range of 30 to 78 years old are still present in significant percentage.  Finally, we show a comparison between our race distribution and the USA race distribution \cite{US_race_gov, US_race}, showing an improvement with the purpose of generalizing CV models and mitigating potential biases \cite{racebias_no, racebias_yes}.   While the USA federal census does not consider Latino/Hispanic as a race, and distribute their classification as an ethnicity distributed across races \cite{US_race_gov}, we have incorporated it as a race in alignment with its recognition as a separate class by current computer vision models \cite{deepface}.  Lastly, although the non-rigid aspect of our routine creates uncertainty about specific levels of our visibility label, it allows the actor to provide data using more natural behaviour, adding fidelity to the actor's performance.

\subsection{Occluded}
\begin{figure}[t]
  \centering
  \includegraphics[trim=0cm .2cm .8cm .8cm, clip, width=.9\linewidth]{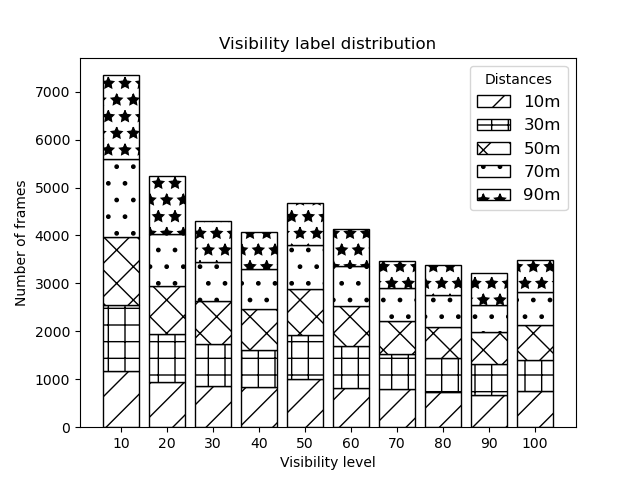}\vspace{-0.2cm}
  \caption{Distribution of the visibility label across the 42,825 manually annotated frames.}\vspace{-0.5cm}
  \label{fig:visibility_dist}
\end{figure}

NOMAD provides the data needed to face occluded persons' detection during high-pressure, life-or-death emergencies. It labels each bounding box with the degree of visibility on 10 levels, providing a representative number of frames at each level as shown in \cref{fig:visibility_dist}. The higher amount of frames at lower visibilities responds to the manual selection process as well as to the increasing  annotation difficulty and additional sources of occlusion at further distances.  

\subsection{Multi-scale}
Effective collaboration between sUAS and humans aims to exploit each of their individual strengths.  sUAS have the ability to quickly scan large areas from greater altitudes, or to provide a focused close-up view of the target.  NOMAD provides five different aerial distances, supporting both generalized models or models specialized for each distance. Table S3 from the Supplemental Material shows the expected minimum Ground Sampling Distance (GSD) for the five different distances, assuming that the actor is on the camera optical axis \cite{gsd_offnadir}. This is not always true as the actors moved to perform their routine and the camera gimbal position was often set to avoid potential areas of non-participants or areas outside the filming premises. This moved the actors away from the camera optical axis, increasing their GSD, and decreasing the number of pixels representing them, as well as creating a non-fixed pitch and adding real variance.

\subsection{Computer Vision Uses}
 The characteristics of NOMAD provide an environment to improve emergency response  in four main areas of CV:\vspace{-3pt}
 \begin{itemize}
    \itemsep=-.3em
     \item Occlusion benchmark: the effort of NOMAD's ten levels of visibility aims to provide a new benchmark dataset to assess the research community's improvements on person detection under occlusion, a previously under-explored factor in aerial datasets.
     \item Person detection: search and rescue scenarios in remote areas tend to search for \textit{any} person (i.e., blind search).  The demographic and environmental diversity provided by NOMAD, as well as its multi-scale component, supported by the bounding boxes' annotations, can be leveraged to improve this general CV task.
     \item Person re-identification: additional to \textit{blind search}, descriptions of the searched person translate the detection task to a CV re-identification problem, especially on crowded scenarios.  NOMAD provides a rich metadata and a reference view of every actor to support re-identification tasks from aerial views.
     \item Person tracking: in many emergency response scenarios, the aim is to detect and then track. Due to the strategic selection of manually labelled keyframes, NOMAD allows the assessment of tracking techniques, following the actor's key movements and changes of direction throughout their full routine. 
 \end{itemize}

\section{Computer Vision Models metrics}
\label{sec:CVmodels}

To demonstrate the use of NOMAD for benchmarking CV models at varying levels of occlusion, we compared the performance of three state-of-the-art CV models. Our first model was YOLOv8 from Ultralytics \cite{yolov8_ultralytics}, representing the most recent upgrade to the YOLO family. YOLOv8 supports real-time detection with limited computational and memory resources, matching the requirements for sUAS-based aerial detection.  Additionally, we selected a FasterRCNN and a RetinaNet model from the Detectron2 library \cite{detectron2}. The specific versions tested are YOLOv8l, FasterRCNN-R101-FPN, and RetinaNet-R101-FPN, with a reported mAP@0.5:0.95 of 52.9, 42.0 and 40.4 on the COCO benchmark, respectively. Available models of YOLOv8x and FasterRCNN-X101-FPN provide higher mAP, nevertheless, the latency of these models increases substantially compared to the gained mAP \cite{yolov8_ultralytics, detectron2modelzoo}.

\begin{figure*}
  \centering
  \begin{subfigure}{.328\linewidth}
    \includegraphics[width=\linewidth, trim={.75cm .2cm 1.3cm .2cm}, clip]{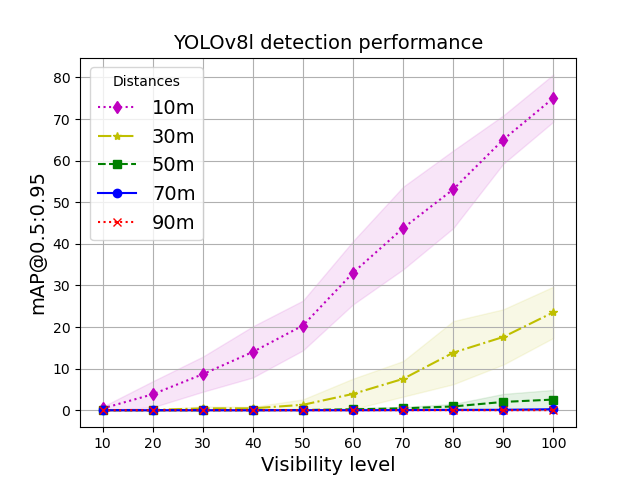}
    \caption{}
    \label{fig:yolo}
  \end{subfigure}
  \begin{subfigure}{.328\linewidth}
    \includegraphics[width=\linewidth, trim={.75cm .2cm 1.3cm .2cm}, clip]{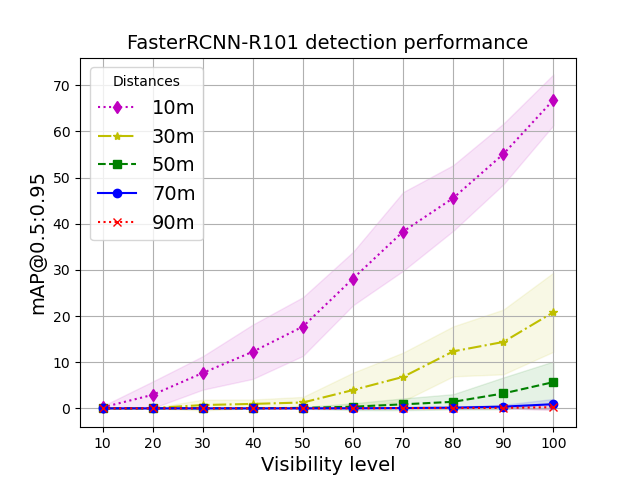}
    \caption{}
    \label{fig:frcnn}
  \end{subfigure}
  \begin{subfigure}{.328\linewidth}
    \includegraphics[width=\linewidth, trim={.75cm .2cm 1.3cm .2cm}, clip]{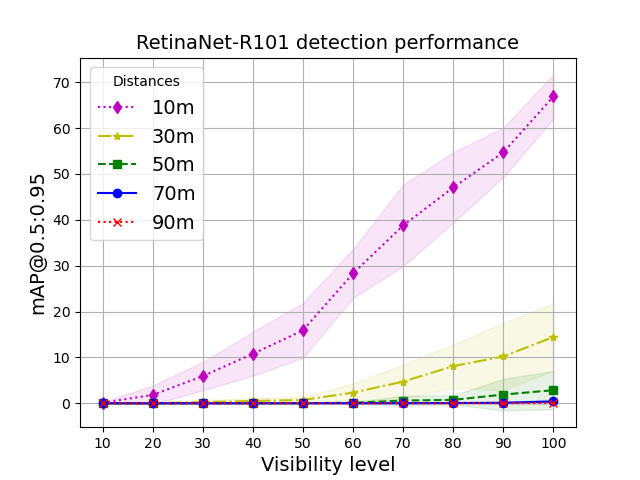}
    \caption{}
    \label{fig:retinanet}
  \end{subfigure}
  \vspace{-0.2cm}
  \caption{Performance across different levels of occlusion of (a) YOLOv8l's, (b) FasterRCNN-R101's, (c) RetinaNet-R101's pretrained weights when tested on NOMAD, with the task of person detection. Oclussion increases as the level of visibility decreases, therefore, mAP scores fall drastically as we increase in distance and occlusion. The higher performance of the models at the closest distance is expected as the data resembles ground view from COCO training data, nevertheless, mAP scores fall significantly as we increase in occlusion even for the closest distance, calling for improvement and robustness of models against occlusion in aerial views.} \vspace{-0.5cm}
  \label{fig:models}
\end{figure*}

\begin{figure}
  \centering
  \begin{subfigure}{.45\linewidth}
    \includegraphics[trim=0cm 1cm 0cm 2cm, clip,width=0.8\linewidth]{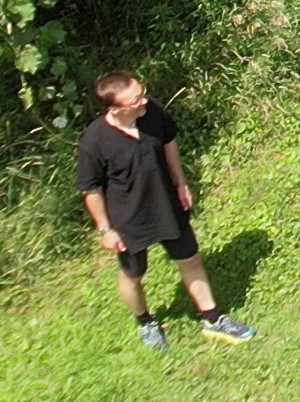}
    \caption{}
    \label{fig:sam-a}
  \end{subfigure}
  \begin{subfigure}{.45\linewidth}
    \includegraphics[trim=0cm 1cm 0cm 2cm, clip,width=0.8\linewidth]{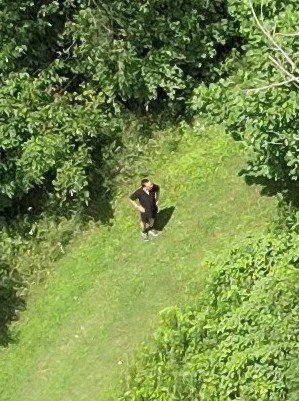}
    \caption{}
    \label{fig:sam-b}
  \end{subfigure}
  \begin{subfigure}{.45\linewidth}
    \includegraphics[trim=0cm 1cm 0cm 2cm, clip,width=0.8\linewidth]{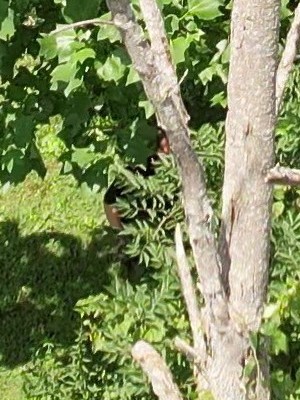}
    \caption{}
    \label{fig:sam-c}
  \end{subfigure}
  \begin{subfigure}{.45\linewidth}
    \includegraphics[trim=0cm 1cm 0cm 2cm, clip,width=0.8\linewidth]{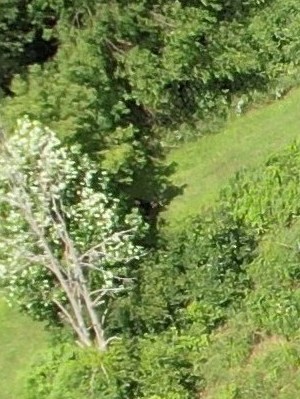}
    \caption{}
    \label{fig:sam-d}
  \end{subfigure}\vspace{-0.2cm}
  \caption{Test samples. (a) Easy sample at 10m and 100 visibility. (b) Medium level difficulty sample (50m, 100 visibility). (c) Hard sample due to heavy occlusion (10m, 30 visibility). (d) Hard sample due to high distance and light occlusion (90m, 90 visibility).}\vspace{-0.35cm}
  \label{fig:samples}
\end{figure}

For evaluation purposes, 10 folds of 10 actors each were randomly created, with a constant seed for reproducibility across models.  From each fold, 50 tests were performed, one per each distance-visibility (5 distances, 10 visibilities), with the results of these 50 tests being averaged across the 10 folds. \Cref{fig:models} shows the averaged mAP@0.5:0.95 score and standard deviation of the CV models against different levels of occlusion and distances. We can observe that YOLOv8l performs better on the closest distance than the other models, a result supported by the higher initial mAP reported; nevertheless all models suffer from critical degradation as the distance increases. This behaviour is expected as their training data is focused on ground views rather than aerial ones.  Finally, although this degradation is expected to be mitigated by fine-tuning the models with aerial data, the results emphasize the degradation problem that occlusion represents, for even though the models achieve decent scores at the closest distance with full visibility, the mAP values drastically drop as the occlusion increases. The usefulness of NOMAD to the research community can be justified from the previous baseline by three reasons: (1) NOMAD is built with a real-world variance imagery (Natural), making it a fair benchmark towards emergency response scenarios; (2) occlusion on person detection can be assessed thanks to the granularity of our visibility label; (3) the multi-scale characteristic allows occlusion to be assessed also across different distances, key to the improvement of aerial detection on sUAS. To exemplify the difficulty of person detection on emergency response scenarios, \cref{fig:samples} shows imagery from our tests with increasing difficulty, due to distance and occlusion.

\section{Future Work and Conclusions}
\label{sec:FutureWorkConclusions}

The structure and characteristics of NOMAD offer many opportunities for improvements in aerial human detection, recognition and tracking, especially the following:
\vspace{-0.05cm}
\begin{itemize}
    \itemsep=-.3em
    \item Detection under occlusion:  NOMAD allows us to explore and understand the limits of detection under occluded views, with future work focusing on improving CV models' performance by exploiting human psychophysical metrics and temporal information.
    \item Person re-identification: Addressing emergency response scenarios, future work will focus on improving person re-identification through leveraging software architectures that support hybrid onboard/offboard solutions and integrate the human into the loop.
    \item Real-world deployment: We have found through our own experiences in deploying CV on sUAS that there are major additional challenges and some degradation in results. Using the models trained on NOMAD, we will deploy and evaluate occlusion-ready CV models on physical sUAS.
\end{itemize}

In conclusion, as indicated by the results of our baseline evaluation, occlusion represents a non-trivial challenge that remains to be tackled.  NOMAD's characteristics of Natural, Occluded, Multi-scale aerial views, provide a new benchmark dataset for tackling this challenge, and can serve as the next step in improving the accuracy of aerial search and detection for emergency response.

\section{Acknowledgements}
\label{sec:Acknowledgements} 
The work described in this paper was supported by the USA National Science Foundation under grant CNS-1931962.
In addition, we thank all participants and annotators, as well as academic institutions, governmental agencies, private owners and company authorized supervisors for providing locations' accessibility. We also thank family and friends for their logistics support.

\balance
{\small
\bibliographystyle{ieee_fullname}
\bibliography{nomad}
}
\newpage
\onecolumn
\appendix
\section{Supplemental tables}

\begin{table}[H]
  \centering
  \begin{tabularx}{\linewidth}{lll}
    \toprule
    Category & Descriptor & Values \\
    \midrule
    Demographic & Age & Years integer.\\
     & Weight & Pounds integer.\\
     & Height & Inches integer.\\
     & Gender & Male, Female\\
     & Race & White Caucasian, Latino Hispanic, African Descent, Asian, South Asian, Middle Eastern, \\ 
     && Pacific Islander.\\
     & Clothes & Including no more than three dominant clothes, described by\\
     && its relative location to the body (Upper, Lower, Both), its color (see color list),\\
     && a pattern (Horizontal Stripes, Vertical Stripes, Both, Plain, Random), and a figure. \\
     & Hat & Any form of head covering, described by color.\\
     & Gloves & Described by color.\\
     & Shoes & Described by color and type.  Types are categorized in 6 classes, by their coverage \\
     && and formality: Bare, Open Informal (\eg sandals), Open Formal (\eg heels), \\
     && Close Informal (\eg tennis), Close Formal (\eg Oxford), and Boots.  \\
     & Facial Hair & Mustache, Beard.  \\
     & Hair & Described by color and length (Bald, Short, Medium, Long).\\
    \midrule
    Environmental & Date & MM:DD format.\\
     & Time & Integer, in 24hr format.\\
     & Location & Lake, Forest, Field, Golf Course, Quarry, Farm, School, School (Nature), Paintball.\\
     & Weather & Temperature in Fahrenheit, wind speed in MPH, and a word descriptor (\eg "Sunny").\\
     & Video & 5.4k video resolution, 30 fps, 5472 by 3078 pixels' frames, and focal length of 8 mm.\\
     & EV & String of 6 values, in increasing order of altitude and starting from the reference.\\
    \bottomrule
  \end{tabularx}
  \caption{Metadata information attached to every actor. Demographic category includes all descriptors that could aid in identification tasks, including outfit description.}
  \label{tab:metadata}
\end{table}

\begin{table}[H]
  \centering
  \begin{tabularx}{\linewidth}{ll}
    \toprule
    Descriptors & Colors \\
    \midrule
    Clothes, Hat, Gloves, Shoes & Red, Orange, Yellow, Green, Cyan, Blue, Purple, Pink, White, Gray, Black, Brown.\\
    Hair & Black, Blond, Sandy, Brown, Gray, White, Red, Green, Blue, Pink, Other.\\
    \bottomrule
  \end{tabularx}
  \caption{List of colors associated to different metadata descriptors.}
  \label{tab:colors}
\end{table}

\begin{table}[H]
  \centering
  \begin{tabular}{lll}
    \toprule
    Distance & GSD [mm/pixel] & Face area [pixels]\\
    \midrule
    10 & 4.242 & 1981 \\
    30 & 12.729 & 220 \\
    50 & 21.213 & 79 \\
    70 & 29.697 & 40 \\
    90 & 38.184 & 24 \\
    \bottomrule
  \end{tabular}
  \caption{Ground Sampling Distance (GSD) values at camera optical axis position, as well as the amount of pixels to cover the area of a person's face assuming a rectangle area of 155mm by 230mm.}
  \label{tab:gsd}
\end{table}

\end{document}